\documentstyle[aclap]{article}

\newcommand{\avmplus}[1]{{\setlength{\arraycolsep}{0.8mm}
                       \renewcommand{\arraystretch}{1.2}
                       \left[
                       \begin{array}{l}
                       #1 \\
                       \end{array}
                       \right]
                    }}

\newcommand{\att}[1]{{\mbox{\normalsize {\bf #1}}}}
\newcommand{\attval}[2]{{\mbox{\normalsize {\sc #1}}\ =\ {{#2}}}}

\def\GL{$\cal GL$}
\def\TDL{$\cal TDL$}

\title{\vspace{-0.5in}A Lexicon for Underspecified Semantic Tagging}
\author{Paul Buitelaar \\
Dept.~of Computer Science \\ 
Brandeis University \\ 
Waltham, MA 02254-9110, USA \\
{\tt paulb@cs.brandeis.edu}}

\begin{document}
\bibliographystyle{fullname}
\maketitle
\vspace{-0.5in}

\begin{abstract}
The paper defends the notion that semantic tagging should be viewed as more 
than {\em disambiguation} between senses. Instead, semantic tagging should 
be a first step in the interpretation process by assigning each lexical 
item a representation of {\em all} of its systematically related senses, 
from which further semantic processing steps can derive discourse dependent 
interpretations. This leads to a new type of semantic lexicon ({\sc CoreLex}) 
that supports {\em underspecified semantic tagging} through a design based on 
{\em systematic polysemous classes} and a class-based acquisition of lexical 
knowledge for specific domains.
\end{abstract}

\section{Underspecified semantic tagging}

Semantic tagging has mostly been considered as nothing more than 
{\em disambiguation} to be performed along the same lines as part-of-speech 
tagging: given {\em n} lexical items each with {\em m} senses apply 
linguistic heuristics and/or statistical measures to pick the most likely 
sense for each lexical item (see eg: \cite{yar:roget} \cite{stev&wilks}). 

\vspace{0.1in}

\noindent
I do not believe this to be the right approach because it blurs the 
distinction between `related' ({\em systematic polysemy}) and `unrelated' 
senses ({\em homonymy} : {\sf bank - bank}). Although homonyms need to be 
tagged with a disambiguated sense, this is not necessarily so in the case 
of systematic polysemy. There are two reasons for this that I will discuss 
briefly here.  

\vspace{0.1in}

\noindent
First, the problem of multiple reference. Consider this example from the 
{\sc brown} corpus:

\vspace{0.1in}

\begin{quote}
{\bf [A \underline{long} book \underline{heavily weighted} with 
\underline{military technicalities}]$_{NP}$}, {\sf in this edition it is 
neither so long nor so technical as it was originally.}
\end{quote}

\vspace{0.3in}

\noindent
The discourse marker ({\sf it}) refers back to an NP that expresses more 
than one interpretation at the same time. The head of the NP ({\sf book}) 
has a number of systematically related senses that are being expressed 
simultaneously. The meaning of {\sf book} in this sentence cannot be 
disambiguated between the number of interpretations that are implied: the 
informational content of the book ({\bf military technicalities}), its 
physical appearance ({\bf heavily weighted}) and the events that are 
involved in its construction and use ({\bf long}).

\vspace{0.1in}

\noindent
The example illustrates the fact that disambiguation between {\em related} 
senses is not always possible, which leads to the further question if a 
discrete distinction between such senses is desirable at all. A number of 
researchers have answered this question negatively (see eg: 
\cite{pus:book} \cite{kill:phd}). Consider these examples from 
{\sc brown}:

\vspace{0.1in}

\begin{tabular}{lll}
(1) & {\bf fast} & {\sf run-up (of the stock)} \\
(2) & {\bf fast} & {\sf action (by the city government)} \\
(3) & {\bf fast} & {\sf footwork (by Washington)} \\
(4) & {\bf fast} & {\sf weight gaining} \\
(5) & {\bf fast} & {\sf condition (of the track)} \\
(6) & {\bf fast} & {\sf response time} \\
(7) & {\bf fast} & {\sf people}\\
(8) & {\bf fast} & {\sf ball} \\
\end{tabular}

\vspace{0.1in}

\noindent
Each use of the adjective `{\sf fast}' in these examples has a slightly 
different interpretation that could be captured in a number of senses, 
reflecting the different syntactic and semantic patterns. For instance:

\vspace{0.1in}

\begin{tabular}{lll}
1. & {\bf `a fast action'} & (1, 2, 3, 4) \\
2. & {\bf `a fast state of affairs'} & (5, 6) \\
3. & {\bf `a fast object'} & (7, 8) \\
\end{tabular}

\vspace{0.1in}

\noindent
On the other hand all of the interpretations have something in common 
also, namely the idea of `speed'. It seems therefore useful to 
{\em underspecify} the lexical meaning of `{\sf fast}' to a representation 
that captures this primary semantic aspect and gives a general structure 
for its combination with other lexical items, both locally (in 
compositional semantics) and globally (in discourse structure).

\vspace{0.1in}

\noindent
Both the {\em multiple reference} and the {\em sense enumeration} problem 
show that lexical items mostly have an indefinite number of related but 
highly discourse dependent interpretations, between which cannot be 
distinguished by semantic tagging alone. Instead, semantic tagging should 
be a first step in the interpretation process by assigning each lexical 
item a representation of all of its systematically related `senses'. 
Further semantic processing steps derive discourse dependent
interpretations from this representation. Semantic tags are therefore more 
like {\em pointers} to complex knowledge representations, which can be 
seen as {\em underspecified} lexical meanings.

\section{{\sc CoreLex}: A Semantic Lexicon with Systematic Polysemous 
Classes} 
\label{CORE}

In this section I describe the structure and content of a lexicon 
({\sc CoreLex}) that builds on the assumptions about lexical semantics and 
discourse outlined above. More specifically, it is to be `structured in 
such a way that it reflects the lexical semantics of a language in 
systematic and predictable ways' \cite{pusbogujohn}. This assumption is 
fundamentally different from the design philosophies behind existing 
lexical semantic resources like {\sc WordNet} that do not account for any 
regularities between senses. For instance, {\sc WordNet} assigns to the 
noun {\sf book} the following senses: 

\begin{figure}[ht]
 \framebox[3.1in]{
\begin{tabular}{l}
\\
\bf{publication} \\
\bf{product, production} \\
\bf{fact} \\
\bf{dramatic\_composition, dramatic\_work} \\
\bf{record} \\
\bf{section, subdivision} \\
\bf{journal} \\
\\
\end{tabular}
  }
\caption{{\sc WordNet} senses for the noun {\sf book}}
\label{book}
\end{figure}

\noindent
At the top of the {\sc WordNet} hierarchy these seven senses can be 
reduced to two unrelated `basic senses': the content that is being 
communicated ({\bf communication}) and the medium of communication 
({\bf artifact}). More accurately, {\sf book} should be assigned a 
qualia structure which implies both of these interpretations and 
connects them to each of the more specific senses that {\sc WordNet} 
assigns: that is, {\bf facts, drama} and a  {\bf journal} can be 
{\em part-of} the content of a book; a {\bf section} is {\em part-of} 
both the content and the medium; {\bf publication, production} and 
{\bf recording} are all {\em events} in which both the content and 
the medium aspects of a book can be involved.

\vspace{0.1in}

\noindent
An important advantage of the {\sc CoreLex} approach is more 
consistency among the assignments of lexical semantic structure. 
Consider the senses that {\sc WordNet} assigns to {\sf door, gate} 
and {\sf window}:

\begin{figure}[ht]
 \framebox[3.1in]{
\begin{tabular}{ll}
\\
\sf{door} \\
\\
\bf{movable\_barrier} & $\leadsto$ artifact \\
\bf{entrance} & $\leadsto$ opening \\
\bf{access} & $\leadsto$ cognition, knowledge  \\
\bf{house} & $\leadsto$ ?? \\
\bf{room} & $\leadsto$ ?? \\
\\
\\
\sf{gate} \\
\\
\bf{movable\_barrier} & $\leadsto$ artifact \\
\bf{computer\_circuit} & $\leadsto$ opening \\
\bf{gross\_income} & $\leadsto$ opening \\
\\
\\
\sf{window} \\
\\
\bf{opening} & $\leadsto$ opening \\
\bf{panel}  & $\leadsto$ artifact \\
\bf{display} & $\leadsto$ cognition, knowledge  \\
\\
\end{tabular}
  }
\caption{{\sc WordNet} senses for the nouns {\sf door}, \\ 
{\sf window} and {\sf gate}}
\label{door,window,gate}
\end{figure}

\noindent
Obviously these are similar words, something which is not expressed 
in the {\sc WordNet} sense assignments. In the {\sc CoreLex} 
approach, these nouns are given the same semantic type, which is 
{\em underspecified} for any specific `sense' but assigns them 
consistently with the same {\em basic} lexical semantic structure 
that expresses the regularities between all of their interpretations. 

\vspace{0.1in}

\noindent
However, despite its shortcomings {\sc WordNet} is a vast resource 
of lexical semantic knowledge that can be mined, restructured and 
extended, which makes it a good starting point for the construction 
of {\sc CoreLex}. The next sections describe how systematic 
polysemous classes and underspecified semantic types can be derived 
from {\sc WordNet}. In this paper I only consider classes of 
{\em nouns}, but the process described here can also be applied to 
other parts of speech.

\subsection{Systematic polysemous classes}

We can arrive at classes of systematically polysemous lexical items by 
investigating which items share the same senses and are thus polysemous 
in the same way. This comparison is done at the top levels of the 
{\sc WordNet} hierarchy. {\sc WordNet} does not have an explicit level 
structure, but for the purpose of this research one can distinguish a 
set of 32 `basic senses' that partly coincides with, but is not based 
directly on {\sc WordNet}'s list of 26 `top types': 

\begin{quote}
{\bf act ({\tt act}), agent ({\tt agt}), animal ({\tt anm}), artifact 
({\tt art}), attribute ({\tt atr}), blunder ({\tt bln}), cell ({\tt cel}), 
chemical ({\tt chm}), communication ({\tt com}), event ({\tt evt}), food 
({\tt fod}), form ({\tt frm}), group\_biological ({\tt grb}), group 
({\tt grp}), group\_social ({\tt grs}), human ({\tt hum}), linear\_measure 
({\tt lme}), location ({\tt loc}), location\_geographical ({\tt log}), 
measure ({\tt mea}), natural\_object ({\tt nat}), phenomenon ({\tt phm}), 
plant ({\tt plt}), possession ({\tt pos}), part ({\tt prt}), psychological 
({\tt psy}), quantity\_definite ({\tt qud}), quantity\_indefinite 
({\tt qui}), relation ({\tt rel}), space ({\tt spc}), state ({\tt sta}), 
time ({\tt tme})}\\
\end{quote}

\noindent
Figure \ref{WordNet-Brown} shows their distribution among noun stems in 
the {\sc brown} corpus. For instance there are 2550 different noun stems 
(with 49,824 instances) that have each 2 out of the 32 `basic senses' 
assigned to them in 238 different combinations (a subset of $32^{2}$ = 
1024 possible combinations).

\begin{figure}[ht]
 \framebox[3.1in]{
  \begin{tabular}{rrrr}
\\
   senses & comb's & stems & instances \\
\\
 2 & 238 & 2550 & 49824 \\ 
 3 & 379 &  936 & 35608 \\ 
 4 & 268 &  347 & 22543 \\ 
 5 & 148 &  154 & 15345 \\
 6 &  52 &   52 &  5915 \\
 7 &  27 &   27 &  5073 \\
 8 &  10 &   10 &  3273 \\
 9 &   3 &    3 &  1450 \\
10 &   1 &    1 &   483 \\
11 &   2 &    2 &   959 \\
12 &   1 &    1 &   441 \\
   & \_\_\_\_\_  & \_\_\_\_\_\_\_\_\_\_ & \_\_\_\_\_\_\_\_\_\_ \\
   &        1161 &	          10797 &               140914 \\
\\
  \end{tabular}
  }
\caption{Polysemy of nouns in {\sc brown}}
\label{WordNet-Brown}
\end{figure}

\noindent
We now reduce all of {\sc WordNet}'s sense assignments to these basic 
senses. For instance, the seven different senses that {\sc WordNet} 
assigns to the lexical item {\sf book} (see Figure \ref{book} above) can 
be reduced to the two basic senses: {\tt `art com'}. We do this for each 
lexical item and then group them into classes according to their 
assignments.

\vspace{0.1in}

\noindent
{}From these one can filter out those classes that have only one member 
because they obviously do not represent a systematically polysemous class. 
The lexical items in those classes have a highly idiosyncratic behavior 
and are most likely homonyms. This leaves a set of 442 polysemous classes, 
of which Figure \ref{Polysemous classes} gives a selection:

\begin{figure}[ht]
 \framebox[3.1in]{
\begin{tabular}{ll}
\\
{\tt act art evt rel} & {\sf click modification reverse} \\
{\tt act art log}     & {\sf berth habitation mooring} \\
{\tt act evt nat}     & {\sf ascent climb} \\
{\tt chm sta}	      & {\sf grease ptomaine} \\
{\tt com prt}	      & {\sf appendix brickbat index} \\
{\tt frm sta}	      & {\sf solid vacancy void} \\
{\tt lme qud}	      & {\sf em fathom fthm inch mil} \\
{\tt loc psy}	      & {\sf bourn bourne demarcation} \\
		      & {\sf fairyland rubicon trend vertex} \\
{\tt log pos sta}     & {\sf barony province} \\
{\tt phm pos}         & {\sf accretion usance wastage} \\ 
{\tt rel sta}         & {\sf baronetcy connectedness} \\
		      & {\sf context efficiency inclusion} \\ 	
                      & {\sf liquid relationship} \\ 	
\\
\end{tabular}
  }
\caption{A selection of polysemous classes}
\label{Polysemous classes}
\end{figure}

\noindent
Not all of the 442 classes are systematically polysemous. Consider for 
example the following classes: 

\begin{figure}[ht]
 \framebox[3.1in]{
\begin{tabular}{ll}
\\
{\tt act anm art}	  & {\sf drill ruff solitaire stud} \\
{\tt act log}	          & {\sf bolivia caliphate charleston} \\
			  & {\sf chicago clearing emirate michigan} \\
		          & {\sf prefecture repair santiago wheeling} \\
{\tt act plt}	          & {\sf chess grapevine rape} \\
{\tt art fod loc}	  & {\sf pike port} \\ 
{\tt chm psy}   	  & {\sf complex incense}  \\ 
{\tt fod hum plt}	  & {\sf mandarin sage swede} \\ 
\\
\end{tabular}
  }
\caption{A selection of {\em ambiguous} classes}
\label{Ambiguous classes}
\end{figure}

\noindent
Some of these classes are collections of homonyms that are {\em ambiguous} 
in similar ways, but do not lead to any kind of predictable polysemous 
behavior, for instance the class {\tt `act anm art'} with the lexical 
items: {\sf drill ruff solitaire stud}. Other classes consist of both 
homonyms and systematically polysemous lexical items like the class 
{\tt act log}, which includes {\sf caliphate, clearing, emirate, 
prefecture, repair, wheeling} vs. {\sf bolivia, charleston, chicago, 
michigan}. Whereas the first group of nouns express two separated but 
related meanings (the {\bf act} of {\sf clearing, repair,} etc. takes place 
at a certain {\bf location}), the second group expresses two meanings that 
are not related (the {\sf charleston} dance which was named after the town 
by the same name).

\vspace{0.1in}

\noindent
The {\em ambiguous} classes need to be removed altogether, while the ones 
with mixed {\em ambiguous} and {\em polysemous} lexical items are to be 
weeded out carefully.

\subsection{Underspecified semantic types}
\label{UST}

The next step in the research is to organize the remaining classes into 
knowledge representations that relate their senses to each other. These 
representations are based on  Generative Lexicon theory (\GL), using 
{\em qualia roles} and {\em (dotted) types} \cite{pus:book}. 

\vspace{0.1in}

\noindent
Qualia roles distinguish different semantic aspects: {\sc formal} indicates 
semantic type; {\sc constitutive} {\em part-whole} information; 
{\sc agentive} and {\sc telic} associated events (the first dealing with 
the {\em origin} of the object, the second with its {\em purpose}). Each 
role is typed to a specific class of lexical items. Types are either simple 
({\bf human, artifact,...}) or complex (e.g., 
{\bf information$\bullet$physical}). Complex types are called 
{\em dotted types} after the `dots' that are used as type constructors. 
Here I introduce two kinds of dots: 

\begin{quote}
Closed dots `$\bullet$' connect systematically related types that are 
always interpreted simultaneously. 
\end{quote}

\begin{quote}
Open dots `$\circ$' connect systematically related types that are not 
(normally) interpreted simultaneously. 
\end{quote}

\noindent
Both `$\sigma$$\bullet$$\tau$' and `$\sigma$$\circ$$\tau$' denote sets of 
{\em pairs} of objects $\langle a, b \rangle$, $a$ an object of type 
$\sigma$ and $b$ an object of type $\tau$. A condition $a${\bf R}$b$ 
restricts this set of pairs to only those for which some relation {\bf R} 
holds, where {\bf R} denotes a subset of the Cartesian product of the sets 
of type $\sigma$ objects and type $\tau$ objects. 

\vspace{0.1in}

\noindent
The difference between types `$\sigma$$\bullet$$\tau$' and 
`$\sigma$$\circ$$\tau$' is in the nature of the objects they denote. The 
type `$\sigma$$\bullet$$\tau$' denotes sets of pairs of objects where each 
pair behaves as a {\em complex} object in discourse structure. For instance, 
the pairs of objects that are introduced by the type 
{\bf information$\bullet$physical} ({\sf book, journal, scoreboard, ...}) 
are addressed as the complex objects $\langle${\bf x:information}, 
{\bf y:physical}$\rangle$ in discourse. On the other hand, the type 
`$\sigma$$\circ$$\tau$' denotes simply a set of pairs of objects that do 
not occur together in discourse structure. For instance, the pairs of 
objects that are introduced by the type {\bf form$\bullet$artifact} 
({\sf door, gate, window, ...}) are not (normally) addressed simultaneously 
in discourse, rather one side of the object is picked out in a particular 
context. Nevertheless, the pair as a whole remains active during processing.

\vspace{0.1in}

\noindent
The resulting representations can be seen as underspecified lexical meanings 
and are therefore referred to as {\em underspecified semantic types}. 
{\sc CoreLex} currently covers 104 underspecified semantic types. This
section presents a number of examples, for a complete overview see the 
{\sc CoreLex} webpage:

\vspace{0.1in}

{\scriptsize \tt http://www.cs.brandeis.edu/\verb+~+paulb/CoreLex/corelex.html} 

\paragraph{Closed Dots} Consider the underspecified representation for the 
semantic type {\bf act$\bullet$relation}:

\begin{figure}[ht]
$\avmplus{
 \attval{formal}
  {\att{Q:act$\bullet$relation}}\\
 \attval{constitutive}
  {\att{}} \\
  {\att{\hspace{0.1in} X:act $\vee$ Y:relation $\vee$ Z:act$\bullet$relation}}
 \\
 \attval{telic}
  {\att{}} \\
  {\att{\hspace{0.1in} P:event(act$\bullet$relation) $\wedge$ 
act(R$_1$) $\wedge$}} \\
  {\att{\hspace{0.1in} relation(R$_2$,R$_3$)}}
}$
\caption{Representation for type: {\bf act$\bullet$relation}}
\label{acr-type}
\end{figure}

\noindent
The representation introduces a number of objects that are of a certain type. 
The {\sc formal} role introduces an object Q of type 
{\bf act$\bullet$relation}. The {\sc constitutive} introduces objects that 
are in a part-whole relationship with Q. These are either of the same type  
{\bf act$\bullet$relation} or of the simple types {\bf act} or 
{\bf relation}. The {\sc telic} expresses the {\bf event} P that can be 
associated with an object of type {\bf act$\bullet$relation}. For instance, 
the event of {\sf increase} as in {\sf `increasing the communication between 
member states'} implies {\sf `increasing'} both the {\bf act} of 
communicating an object R$_1$ and the communication {\bf relation} between 
two objects R$_2$ and R$_3$. All these objects are introduced on the semantic 
level and correspond to a number of objects that will be realized in syntax. 
However, not all semantic objects will be realized in syntax. (See Section 
\ref{LKA} for more on the syntax-semantics interface.)

\vspace{0.1in}

\noindent
The instances for the type {\bf act$\bullet$relation} are given in Figure 
\ref{acr}, covering three different systematic polysemous classes. We could 
have chosen to include only the instances of the {\tt `act rel'} class, but 
the nouns in the other two classes seem similar enough to describe all of 
them with the same type. 

\begin{figure}[ht]
 \framebox[3.1in]{
\begin{tabular}{ll}
\\
{\tt act evt rel} & {\sf blend competition flux} \\ 
                  & {\sf transformation} \\ 
{\tt act rel}	  & {\sf acceleration communication } \\
                  & {\sf dealings designation discourse gait} \\
                  & {\sf glide likening negation neologism} \\ 
                  & {\sf neology prevention qualifying} \\
                  & {\sf sharing synchronisation} \\ 
                  & {\sf synchronization synchronizing} \\ 
{\tt act rel sta} & {\sf coordination gradation involvement} \\ 
\\
\end{tabular}
  }
\caption{Instances for the type: {\bf act$\bullet$relation}}
\label{acr}
\end{figure}

\paragraph{Open Dots} The type {\bf act$\bullet$relation} describes 
interpretations that can not be separated from each other (the {\bf act} 
and {\bf relation} aspects are intimately connected). The following 
representation for type {\bf animal$\circ$food} describes interpretations 
that can not occur simultaneously but are however related\footnote{
	See the literature on {\em animal grinding}, for instance 
\cite{cobris:lexop}}. 
It therefore uses a `$\circ$' instead of a `$\bullet$' as a type 
constructor:

\begin{figure}[ht]
$\avmplus{
 \attval{formal}
  {\att{Q:animal$\circ$food}}\\
 \attval{constitutive}
  {\att{X:animal $\vee$ Y:food}} \\
 \attval{telic}
  {\att{}} \\
  {\att{\hspace{0.1in} P$_1$:act(R$_1$,animal) $\vee$ 
P$_2$:act(animal,R$_2$)}} \\
  {\att{\hspace{0.1in} $\vee$ P$_3$:act(R$_3$,food)}}
}$
\caption{Representation for type: {\bf animal$\circ$food}}
\end{figure}

\noindent
The instances for this type only cover the class {\tt `anm fod'}. A case 
could be made for including also every instance of the class {\tt `anm'} 
because in principal every animal could be eaten. This is a question of 
how {\em generative} the lexicon should be and if one allows 
{\em overgeneration} of semantic objects. 

\begin{figure}[ht]
 \framebox[3.1in]{
\begin{tabular}{ll}
\\
{\tt anm fod} & {\sf bluepoint capon clam cockle crawdad} \\
              & {\sf crawfish crayfish duckling fowl} \\
              & {\sf grub hen lamb langouste limpet} \\ 
              & {\sf lobster monkfish mussel octopus panfish} \\
              & {\sf partridge pheasant pigeon poultry} \\ 
              & {\sf prawn pullet quail saki scallop} \\
              & {\sf scollop shellfish shrimp snail} \\ 
              & {\sf squid whelk whitebait whitefish winkle} \\ 
\\
\end{tabular}
  }
\caption{Instances for the type: {\bf animal$\circ$food}}
\label{anf}
\end{figure}

\subsection{Homonyms}

{\sc CoreLex} is designed around the idea of systematic polysemous classes 
that exclude homonyms. Traditionally a lot of research in lexical semantics 
has been occupied with the problem of ambiguity in homonyms. Our research 
shows however that homonyms only make up a fraction of the whole of the 
lexicon of a language. Out of the 37,793 noun stems that were derived from 
{\sc WordNet} 1637 are to be viewed as true homonyms because they have two 
or more {\em unrelated} senses, less than 5\%. The remaining 95\% are nouns 
that do have (an indefinite number of) different interpretations, but all 
of these are somehow related and should be inferred from a common knowledge 
representation. These numbers suggest a stronger emphasis in research on 
systematic polysemy and less on homonyms, an approach that is advocated 
here (see also \cite{kill:phd}). 

\vspace{0.1in}

\noindent
In {\sc CoreLex} homonyms are simply assigned two or more underspecified 
semantic types, that need to be disambiguated in a traditional way. There 
is however an added value also here because each disambiguated type can 
generate any number of context dependent interpretations.

\section{Adapting {\sc CoreLex} to Domain Specific Corpora}

The underspecified semantic type that {\sc CoreLex} assigns to a noun 
provides a basic lexical semantic structure that can be seen as the 
class-wide backbone semantic description on top of which specific 
information for each lexical item is to be defined.

\vspace{0.1in}

\noindent
That is, {\sf doors} and {\sf gates} are both artifacts but they have 
different appearances. Gates are typically open constructions, whereas 
doors tend to be solid. This kind of information however is corpus 
specific and therefore needs to be adapted specifically to and on the 
basis of that particular corpus of texts. 

\vspace{0.1in}

\noindent
This process involves a number of consecutive steps that includes the 
probabilistic classification of unknown lexical items: 

\begin{enumerate}
\item Assignment of underspecified semantic tags to those nouns that 
are in {\sc CoreLex}
\item Running class-sensitive patterns over the (partly) tagged corpus
\item
 \begin{enumerate}
 \item Constructing a probabilistic classifier from the data obtained in 
step 2.
 \item Probabilistically tag nouns that are not in {\sc CoreLex} according 
to this classifier
 \end{enumerate}
\item Relating the data obtained in step 2. to one or more qualia roles 
\end{enumerate}

\noindent
Step 1. is trivial, but steps 2. through 4. form a complex process of 
constructing a corpus specific semantic lexicon that is to be used in 
additional processing for knowledge intensive reasoning steps (i.e. 
{\em abduction} \cite{hobbs:abduct}) that would solve metaphoric, 
metonymic and other non-literal use of language.

\subsection{Assignment of {\sc CoreLex} Tags} 

The first step in analyzing a new corpus involves tagging each noun that is 
in {\sc CoreLex} with an underspecified semantic tag. This tag represents 
the following information: a definition of the type of the noun 
({\sc formal}); a definition of types of possible nouns it can stand in a 
part-whole relationship with ({\sc constitutive}); a definition of types of 
possible verbs it can occur with and their argument structures 
({\sc agentive / telic}). {\sc CoreLex} is implemented as a database of 
associative arrays, which allows a fast lookup of this information in 
pattern matching.

\subsection{Class-Sensitive Pattern Matching} 

The pattern matcher runs over corpora that are: part-of-speech tagged using 
a widely used tagger \cite{brill}; stemmed by using an experimental system 
that extends the Porter stemmer, a stemming algorithm widely used in 
information retrieval, with the Celex database on English morphology; 
(partly) semantically tagged using the {\sc CoreLex} set of underspecified 
semantic tags as discussed in the previous section. 

\vspace{0.1in}

\noindent
There are about 30 different patterns that are arranged around the 
{\bf headnoun} of an NP. They cover the following syntactic constructions that 
roughly correspond to a VP, an S, an NP and an NP followed by a PP:

\begin{itemize}
\item {\bf verb-headnoun}
\item {\bf headnoun-verb}
\item {\bf adjective-headnoun}
\item {\bf modifiernoun-headnoun}
\item {\bf headnoun-preposition-headnoun} 
\end{itemize}

\noindent
The patterns assume NP's of the following generic structure\footnote{
	The interpretation of `{\tt *}' and `{\tt ?}' in this section follows 
	that of common usage in regular expressions: `{\tt *}' indicates 0 or 
	more occurrences; `{\tt ?}' indicates 0 or 1 occurrence}:

\vspace{0.1in}

\noindent
{\tt PreDet* Det* Num* (Adj|Name|Noun)* Noun} 
\vspace{0.1in}

\noindent
The heuristics for finding the {\bf headnoun} is then simply to take the 
rightmost noun in the NP, which for English is mostly correct.

\vspace{0.1in}

\noindent
The {\bf verb-headnoun} patterns approach that of a true `verb-obj' analysis 
by including a normalization of passive constructions as follows:

\vspace{0.1in}

\noindent
{\tt [Noun Have? Be Adv? Verb]} $\Rightarrow$ {\tt [Verb Noun]} 

\vspace{0.1in}

\noindent
Similarly, the {\bf headnoun-verb} patterns approach a true `subj-verb' 
analysis. However, because no deep syntactic analysis is performed, the 
patterns can only approximate subjects and objects in this way and I 
therefore do not refer to these patterns as `{\bf subject-verb}' and 
`{\bf verb-object}' respectively.

\vspace{0.1in}

\noindent
The pattern matching is class-sensitive in employing the assigned 
{\sc CoreLex} tag to determine if the application of this pattern is 
appropriate. For instance, one of the {\bf headnoun-preposition-headnoun} 
patterns is the following, that is used to detect {\em part-whole} 
({\sc constitutive}) relations:

\vspace{0.1in}

\noindent
{\tt PreDet* Det* Num* (Adj|Name|Noun)* Noun of PreDet* Det* Num* 
(Adj|Name|Noun)* Noun} 
\vspace{0.1in}

\noindent
Clearly not every syntactic construction that fits this pattern is to be 
interpreted as the expression of a part-whole relation. One of the 
heuristics we therefore use is that the pattern may only apply if both 
head nouns carry the same {\sc CoreLex} tag or if the tag of the second 
head noun subsumes the tag of the first one through a dotted type. That 
is, if the second head noun is of a dotted type and the first is of one 
of its composing types. For instance, `{\sf paragraph}' and 
`{\sf journal}' can be in a {\em part-whole} relation to each other 
because the first is of type {\bf information}, while the second is of 
type {\bf information$\bullet$physical}. Similar heuristics can be 
identified for the application of other patterns. 

\vspace{0.1in}

\noindent
Recall of the patterns (percentage of nouns that are covered) is on 
average, among different corpora ({\sc wsj}, {\sc brown}, {\sc pdgf} -- 
a corpus we constructed for independent purposes from 1000 medical 
abstracts in the {\sc medline} database on Platelet Derived Growth 
Factor -- and {\sc darwin} -- the complete {\em Origin of Species}), 
about 70\% to 80\%. Precision is much harder to measure, but depends 
both on the accuracy of the output of the part-of-speech tagger and on 
the accuracy of class-sensitive heuristics.

\subsection{Probabilistic Classification} 
\label{PC}

The knowledge about the linguistic context of nouns in the corpus that is 
collected by the pattern matcher is now used to classify unknown nouns. 
This involves a similarity measure between the linguistic contexts of 
classes of nouns that are in {\sc CoreLex} and the linguistic context of 
unknown nouns. For this purpose the pattern matcher keeps two separate 
arrays, one that collects knowledge only on {\sc CoreLex} nouns and the 
other collecting knowledge on {\em all} nouns. 

\vspace{0.1in}

\noindent
The classifier uses {\em mutual information} (MI) scores rather than the 
raw frequences of the occurring patterns \cite{church&hanks:mi}. Computing 
MI scores is by now a standard procedure for measuring the co-occurrence 
between objects relative to their overall occurrence. MI is defined in 
general as follows:

\[ I \ (x \ y) \ = \ log_{2} \ \frac{P(x \ y)}{P(x) \ P(y)}\]

\noindent
We can use this definition to derive an estimate of the {\em connectedness} 
between words, in terms of collocations \cite{smadja:xtract}, but also in 
terms of phrases and grammatical relations \cite{hindle}. For instance the 
co-occurrence of verbs and the heads of their NP objects ({\em N}: size of 
the corpus, i.e. the number of stems):  

\[ C_{obj} \ (v \ n) \ = \ log_{2} \ \frac{\frac{f(v \ n)} N}{\frac{f(v)}{N} 
\ \frac{f(n)}{N}}\]

\noindent
All nouns are now classified by running a similarity measure over their MI 
scores and the MI scores of each {\sc CoreLex} class. For this we use the 
{\em Jaccard measure} that compares objects relative to the attributes they 
share \cite{grefenstette}. In our case the `attributes' are the different 
linguistic constructions a noun occurs in: {\bf headnoun-verb}, 
{\bf adjective-headnoun}, {\bf modifiernoun-headnoun}, etc.

\vspace{0.1in}

\noindent
The Jaccard measure is defined as the number of attributes shared by two 
objects divided by the total number of unique attributes shared by both 
objects:

\[\frac{A}{A \ + \ B \ + \ C}\]

\vspace{0.1in}

\hspace{0.2in} $A \ : \ $ attributes shared by both objects

\hspace{0.2in} $B \ : \ $ attributes unique to object 1

\hspace{0.2in} $C \ : \ $ attributes unique to object 2

\vspace{0.1in}

\noindent
The Jaccard scores for each {\sc CoreLex} class are sorted and the class 
with the highest score is assigned to the noun. If the highest score is 
equal to 0, no class is assigned. 

\vspace{0.1in}

\noindent
The classification process is evaluated in terms of precision and recall 
figures, but not directly on the classified unknown nouns, because their 
precision is hard to measure. Rather we compute precision and recall on the 
classification of those nouns that are in CoreLex, because we can check 
their class automatically. The assumption then is that the precision and 
recall figures for the classification of nouns that are known correspond to 
those that are unknown. An additional measure of the effectiveness of the 
classifier is measuring the recall on classification of {\em all} nouns, 
known and unknown. This number seems to correlate with the size of the 
corpus, in larger corpora more nouns are being classified, but not 
necessarily more correctly. Correct classification rather seems to depend on 
the homogeneity of the corpus: if it is written in one style, with one theme 
and so on.

\vspace{0.1in}

\noindent
Recall of the classifier (percentage of all nouns that are classified $>$ 0) 
is on average, among different larger corpora ($>$ 100,000 tokens), about 
80\% to 90\%. Recall on the nouns in {\sc CoreLex} is between 35\% and 55\%, 
while precision is between 20\% and 40\%. The last number is much better on 
smaller corpora (70\% on average). More detailed information about the 
performance of the classifier, matcher and acquisition tool (see below) can 
be obtained from \cite{buit:phd}.

\subsection{Lexical Knowledge Acquisition} 
\label{LKA}

The final step in the process of adapting {\sc CoreLex} to a specific domain
involves the `translation' of observed syntactic patterns into corresponding 
semantic ones and generating a semantic lexicon representing that 
information. 

\vspace{0.1in}

\noindent
There are basically three kinds of semantic patterns that are utilized in a 
{\sc CoreLex} lexicon: hyponymy (sub-supertype information) in the 
{\sc formal} role, meronymy (part-whole information) in the 
{\sc constitutive} role and predicate-argument structure  in the {\sc telic} 
and {\sc agentive} roles. There are no compelling reasons to exclude other 
kinds of information, but for now we base our basic design on \GL \space, 
which only includes these three in its definition of qualia structure. 

\vspace{0.1in}

\noindent
Hyponymic information is acquired through the classification process 
discussed in Sections \ref{UST} and \ref{PC}. Meronymic information is 
obtained through a translation of various VP and PP patterns into `has-part' 
and `part-of' relations. Predicate-argument structure finally, is derived 
from {\bf verb-headnoun} and {\bf headnoun-verb} constructions.

\vspace{0.1in}

\noindent
The semantic lexicon that is generated in such a way comes in two formats: 
\TDL, a {\em Type Description Language} based on typed feature-logic 
\cite{tdl1} \cite{tdl2} and HTML, the markup language for the World Wide Web. 
The first provides a constraint-based formalism that allows {\sc CoreLex} 
lexicons to be used straightforwardly in constraint-based grammars. The 
second format is used to present a generated semantic lexicon as a semantic 
index on a World Wide Web document. We will not elaborate on this further 
because the subject of semantic indexing is out of the scope of this paper, 
but we refer to \cite{pusetal:hype}.

\subsection{An Example: The {\sc pdgf} Lexicon} 

The semantic lexicon we generated for the {\sc pdgf} corpus covers 1830 noun 
stems, spread over 81 {\sc CoreLex} types. For instance, the noun 
{\sf evidence} is of type {\bf communication$\bullet$psychological} and the 
following representation is generated:

\begin{figure}[h]
$\avmplus{
 {\att{\sf evidence}} \\
 \\
 \attval{formal}
  {\att{}} \\
  \hspace{0.1in} 
  {\avmplus{
   \attval{closed}
    {\avmplus{
     \attval{arg1}
      {\att{communication}} \\
     \attval{arg2}
      {\att{psychological}}
    }}
  }} \\
 \attval{constitutive}
  {\att{}} \\
  \hspace{0.1in} 
  {\avmplus{
   \attval{has-part}
    {\avmplus{
     \attval{first}
      {\att{structure}} \\
     \attval{rest}
      {\att{...}}
    }} \\
  }} \\
 \attval{telic}
  {\att{}} \\
  \hspace{0.1in} 
  {\avmplus{
   \attval{first}
    {\avmplus{
    {\att{provide}} \\
     \attval{arg-struct}
      {\att{...}}
    }} \\
   \\
   \attval{rest}
      {\att{...}}
  }}
}$
\caption{Lexical entry for: {\sf evidence}}
\end{figure}

\section{Conclusion}

In this paper I discuss the construction of a new type of semantic lexicon 
that supports  {\em underspecified} semantic tagging. Traditional semantic 
tagging assumes a number of distinct senses for each lexical item between 
which the system should choose. Underspecified semantic tagging however 
assumes no finite lists of senses, but instead tags each lexical item with 
a comprehensive knowledge representation from which a specific 
{\em interpretation} can be {\em constructed}. {\sc CoreLex} provides such 
knowledge representations, and as such it is fundamentally different from 
existing semantic lexicons like {\sc WordNet}. Additionally, it was shown 
that {\sc CoreLex} provides for more consistent assignments of lexical 
semantic structure among classes of lexical items. Finally, the approach 
described above allows one to generate domain specific semantic lexicons 
by enhancing {\sc Corelex} lexical entries with corpus based information.

\end{document}